\documentclass[letterpaper, 10 pt, journal, twoside]{IEEEtran}
\IEEEoverridecommandlockouts

\usepackage{times}
\usepackage{makecell}

\usepackage{etoolbox}
\makeatletter
\patchcmd{\@makecaption}
  {\scshape}
  {}
  {}
  {}
\makeatother

\usepackage{tikz}

\usepackage[numbers]{natbib}
\usepackage{multicol}
\usepackage{multirow}
\usepackage{graphicx}
\usepackage[bookmarks=true]{hyperref}
\usepackage{amsmath}
\usepackage{amssymb}
\usepackage{float}
\usepackage{pifont}
\usepackage{placeins}
\usepackage{mathdots}
\usepackage[caption=false,font=normalsize,labelfont=sf,textfont=sf]{subfig}
\usepackage{xspace}
\usepackage{svg}
\usepackage{rotating}
\usepackage{fontawesome}
\usepackage{pifont}
\usepackage{bm}

\newcommand{\algname}{LiHi-GS\xspace}

\usepackage[table]{xcolor}
\newcommand{\best}[1]{\hspace{-1pt}\cellcolor{red!30}#1\hspace{-1pt}}
\newcommand{\second}[1]{\hspace{-1pt}\cellcolor{orange!30}#1\hspace{-1pt}}
\newcommand{\third}[1]{\hspace{-1pt}\cellcolor{yellow!30}#1\hspace{-1pt}}

\begin{document}


\title{LiHi-GS: \textbf{Li}DAR-Supervised Gaussian Splatting for \textbf{Hi}ghway Driving Scene Reconstruction}

\author{
Pou-Chun Kung$^{1,2,\dagger}$,
Xianling Zhang$^{1}$,
Katherine A. Skinner$^{2}$,
Nikita Jaipuria$^{1}$

\thanks{Manuscript received: April 6, 2025; Revised June 23, 2025; Accepted October 2, 2025.} 
\thanks{This paper was recommended for publication by Editor Pascal Vasseur upon evaluation of the Associate Editor and Reviewers' comments.}
\thanks{$^1$X. Zhang and N. Jaipuria are with Latitude AI. \texttt{\{xzhang, njaipuria\}@lat.ai}.}
\thanks{$^2$P. Kung and K. A. Skinner are with the Department of Robotics, University of Michigan, Ann Arbor, MI 48109. \texttt{\{pckung, kskin\}@umich.edu}.}
\thanks{$^\dagger$ The project was carried out during P. Kung's internship at Latitude AI.}
\thanks{Digital Object Identifier (DOI): see top of this page.}
}

\markboth{IEEE Robotics and Automation Letters. Preprint Version. Accepted October, 2025}{Kung \MakeLowercase{\textit{et al.}}: LiHi-GS: \textbf{Li}DAR-Supervised Gaussian Splatting for \textbf{Hi}ghway Driving Scene Reconstruction}

\maketitle

\begin{figure*}[t!]
    \centering
    \vspace{0.05in}
    \includegraphics[width=1.0\linewidth]{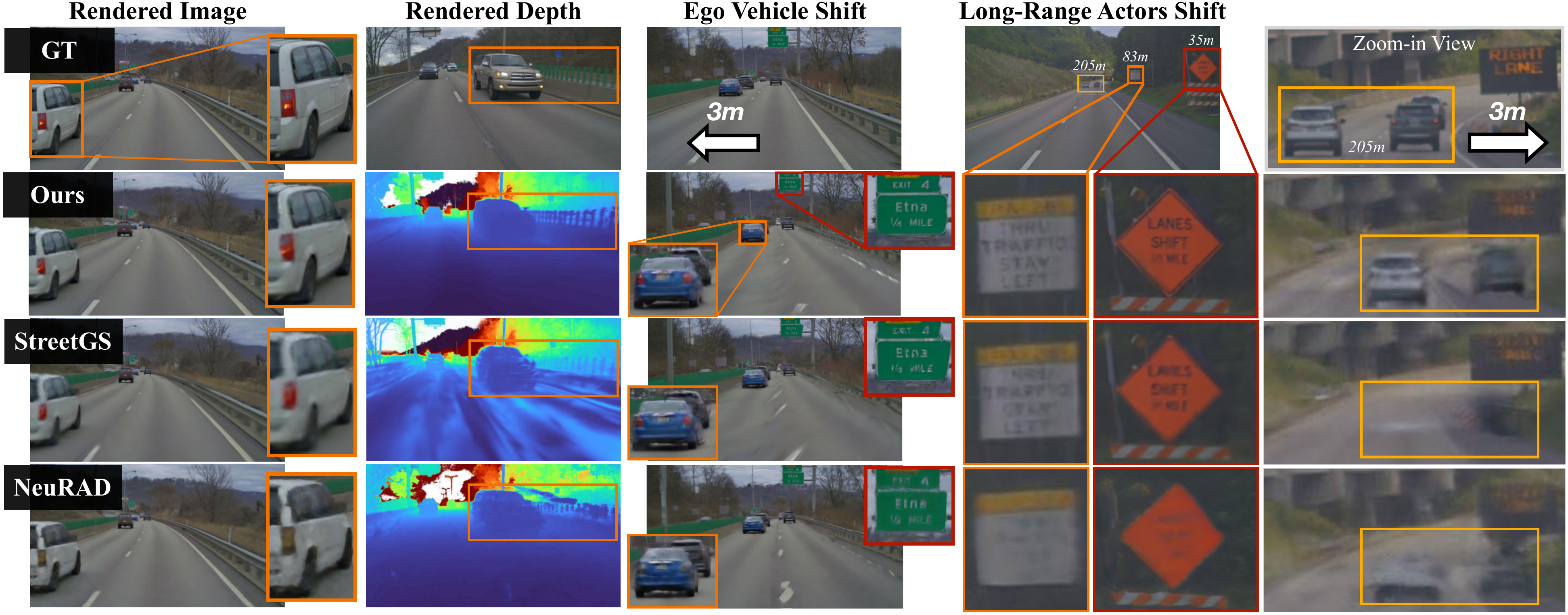}
    \vspace{-0.1in}
    \caption{
    \footnotesize \algname (second row) provides higher quality color and depth renderings for interpolated novel views and for ego/actor shifts compared to state-of-the-art NeRF and GS-based methods~\cite{neurad, streetgs} \algname does particularly well on actor shifts at longer-ranges (205 meters).
    \vspace{-0.25in}
    }
    \label{fig:teaser figure}
\end{figure*}

\begin{abstract}
Photorealistic 3D scene reconstruction plays an important role in autonomous driving, enabling the generation of novel data from existing datasets to simulate safety-critical scenarios and expand training data without additional acquisition costs. Gaussian Splatting (GS) facilitates real-time, photorealistic rendering with an explicit 3D Gaussian representation of the scene, providing faster processing and more intuitive scene editing than the implicit Neural Radiance Fields (NeRFs). While extensive GS research has yielded promising advancements in autonomous driving applications, they overlook two critical aspects. First, existing methods mainly focus on low-speed and feature-rich urban scenes and ignore the fact that highway scenarios play a significant role in autonomous driving. Second, while LiDARs are commonplace in autonomous driving platforms, existing methods learn primarily from images and use LiDAR only for initial estimates or without precise sensor modeling, thus missing out on leveraging the rich depth information LiDAR offers and limiting the ability to synthesize LiDAR data. In this paper, we propose a novel GS method for dynamic scene synthesis and editing with improved scene reconstruction through LiDAR supervision and support for LiDAR rendering. Unlike prior works that are tested mostly on urban datasets, to the best of our knowledge, we are the first to focus on the more challenging and highly relevant highway scenes for autonomous driving, which feature sparse sensor views and monotone backgrounds.
A project page is available at \href{https://umautobots.github.io/lihi_gs}{https://umautobots.github.io/lihi\_gs}.
\end{abstract}


\section{Introduction} \label{introduction}
\IEEEPARstart{W}{hile} there has been a lot of recent progress in semi-supervised and weakly supervised deep learning, in practice, most vision tasks for automated driving still rely on supervised learning and often fail to generalize to unseen scenarios. No matter how big the size of the dataset, capturing long tails is impractical, and neglecting them can have devastating consequences~\cite{teslacrash}. Dataset diversity is thus key to successful real-world deployment. However, data collection and human labeling remain time-intensive and costly.

Synthetic data offers a cost-effective approach to enhance diversity and capture long tails~\cite{zhang2022simbar}. One such source is gaming engine-based simulation (e.g.,\ CARLA~\cite{carla}), which provides perfect annotation for free but lacks realism. More recently, Neural Radiance Fields (NeRFs)~\cite{nerf} have emerged as a popular choice for photorealistic novel view synthesis and scene editing. Prior works have shown promising results in autonomous driving scenarios~\cite{neurad, NSG, unisim, lidar-nerf, NFL, krishnan2023lane}.
However, due to the sampling step in NeRF, these approaches face limitations such as computationally intensive rendering and degraded rendering quality at longer distances~\cite{meganerf}, which are important for highway driving.


In contrast, 3D Gaussian Splatting's (GS)~\cite{3dgs} explicit scene representation enables faster and improved rendering for larger scenes and longer distances, coupled with intuitive scene editing.
Recent works have shown promising results for autonomous driving scenarios~\cite{drivinggs, streetgs, autosplat, HUGS, tclcgs}. 
However, they are limited to urban environments even though autonomous driving applications extend far beyond city streets. In particular, highway scenarios form a major portion of the operating domain for commercial Advanced Driver Assistance Systems (ADAS) and pose distinct challenges for scene reconstruction, such as sparse sensor viewpoints, uniform backgrounds, and repetitive patterns. All of these combined make geometry learning for highway scene reconstruction difficult compared to the feature-rich and lower-speed urban settings.

LiDAR offers dense and precise depth measurements that enhance 3D scene understanding in challenging highway scenarios where camera images suffer from sparse viewpoints and a lack of features.
However, existing works only superficially use LiDAR either for initialization~\cite{autosplat, HUGS} or basic positional alignment~\cite{drivinggs, liv-gaussmap} and, therefore, fail to generalize to LiDAR-sparse regions. More recent approaches attempt to leverage LiDAR depth measurements by projecting point clouds onto camera image planes for depth supervision~\cite{streetgs, omnire, s3gs, letsgo}. 
However, they only make use of the LiDAR measurements that overlap with camera views, make a strong simplifying assumption that the LiDAR sensor is physically close to the cameras, and fail to support LiDAR novel view synthesis. 

To address these limitations, we propose \algname, a GS method with explicit LiDAR sensor modeling. \algname not only enables LiDAR supervision during training, resulting in significantly improved scene geometry learning and novel view image rendering, but it also provides the ability for realistic novel view LiDAR rendering. While prior works have primarily focused on urban close-range scene reconstruction and editing (0-50 meters)~\cite{neurad, streetgs}, we bridge this research gap by conducting comprehensive evaluations on highway scenes with objects at 200 meters and beyond, where the benefits of LiDAR supervision become particularly evident (see Figure~\ref{fig:teaser figure}). Our key contributions are as follows:
\begin{itemize}
   \item {Proposed a differentiable LiDAR rendering model for GS, enabling both LiDAR rendering and supervision.}
   \item {Demonstrated the importance of LiDAR supervision in GS for image and LiDAR novel view synthesis.}
   \item {\algname outperforms state-of-the-art (SOTA) methods in both image and LiDAR synthesis, particularly for view interpolation, ego-view changes, and scene editing tasks.}
   \item {Presented the first comprehensive study on long-range highway scene reconstruction and editing, addressing a crucial yet underdeveloped use case in existing research.}
\end{itemize}

\section{Related Work}
\label{sec:related works}



\begin{figure}[t!]
    \centering
    \includegraphics[width=1.\linewidth]{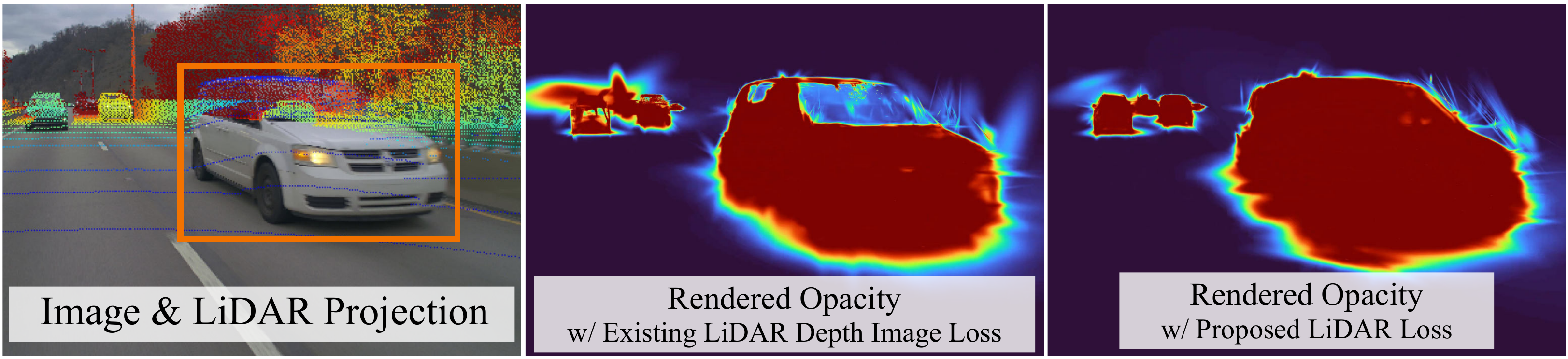}
        \vspace{-0.3in}
        \caption{\footnotesize (Left) Issues with LiDAR projected pseudo-depth supervision, highlighted in the orange box. Points from both near and far distances can map to the same image pixel, resulting in depth ambiguity. In the rendered opacity view (middle), vehicles appear distorted in the case of pseudo-depth supervision, whereas \algname (right) preserves object geometry and integrity.}
        \label{fig:lidar depth image loss}
\end{figure}

\begin{figure*}[t!]
    \centering
    \vspace{0.05in}
    \includegraphics[width=0.95\linewidth]{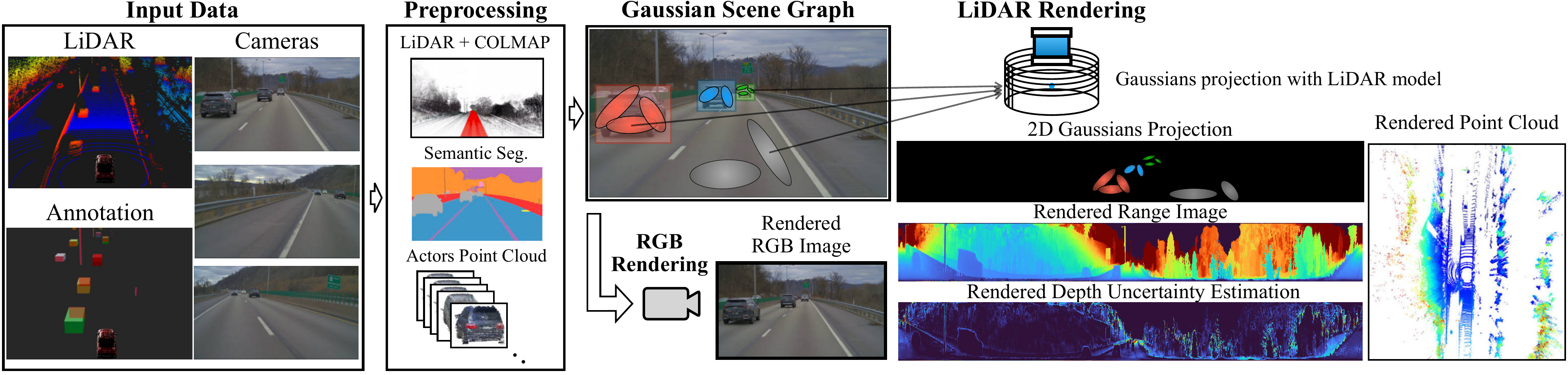}
        \vspace{-0.05in}
        \caption{System overview. \algname takes multiple cameras, LiDAR, and annotated 3D poses as input. During preprocessing, a LiDAR map combined with a COLMAP sparse point cloud is used to initialize the static scene, while LiDAR points aggregated with human-annotated bounding boxes are used to initialize dynamic objects. During training, images and LiDAR scans are rendered separately and compared with the corresponding ground truth images and LiDAR scans for scene reconstruction.}
        \vspace{-0.1in}
        \label{fig:system_diagram}
\end{figure*}

\subsection{NeRF-Based Driving Scene Synthesis} NeRF~\cite{nerf} was originally designed for static room-scaled scenes. Follow-up works like~\cite{blocknerf} extended it to city-scale reconstruction but are also limited to modeling static backgrounds. To model dynamic actors, Neural Scene Graph (NSG)~\cite{NSG} and MARS~\cite{MARS} create a scene graph in which each actor is modeled as an independent NeRF model, and annotated 3D poses are used for scene composition. SUDS~\cite{SUDS} and EmerNeRF~\cite{emernerf} are unsupervised dynamic scene modeling methods that do not require annotated poses. Some studies also estimate ego-vehicle poses~\cite{nerfgs_slam_survey, stereo_gs_slam}.

\subsection{LiDAR-Integrated NeRFs}
Many recent works extend the NeRF formulation for LiDAR rendering~\cite{urf, cloner, loner}. Others focus on LiDAR-only novel view synthesis, like LiDAR-NeRF~\cite{lidar-nerf} and NFL~\cite{NFL}.
DyNFL~\cite{dynNFL} and LiDAR4D~\cite{lidar4d} further extend it to dynamic scenes. UniSim~\cite{unisim} follows NSG-style dynamic scene modeling and supports both image and LiDAR supervision. 
NeuRAD~\cite{neurad} further addresses multiple approximations in UniSim and models camera rolling shutter, LiDAR ray drop, and LiDAR intensity for more realistic novel view synthesis.

\subsection{GS-Based Driving Scene Synthesis}
3DGS~\cite{3dgs} explicitly represents scenes with Gaussians. The Gaussian projection and rasterization expedited training and also enabled real-time rendering. DeformGS~\cite{deformgs} extended 3DGS to reconstruct deformable objects. PVG~\cite{pvg} is one of the first GS methods to reconstruct autonomous driving scenes.
Recent works~\cite{streetgs, drivinggs, autosplat, HUGS} extend the NSG idea to GS and show promising results for driving scenarios. 

\subsection{LiDAR-Integrated GS}
In contrast to NeRFs, LiDAR measurements can be integrated into GS in multiple ways, of which the most common is initializing Gaussians with the LiDAR point cloud as a geometric prior~\cite{autosplat, HUGS}. However, this approach fails to fully integrate LiDAR data into the training pipeline, leading to potential inaccuracies in scene geometry as the model may overfit to camera images used for training. 
To tightly integrate LiDAR measurements into GS training, some studies indirectly supervise the GS model with a LiDAR-supervised NeRF model~\cite{tclcgs} or a LiDAR point cloud map represented by a Gaussian Mixture Model~\cite{ligs}. However, this indirect training fails to provide LiDAR rendering capability.

Alternatively, some works introduce an additional loss to encourage Gaussian centers to align with the LiDAR point cloud~\cite{drivinggs, liv-gaussmap}. This approach, however, assumes that the LiDAR point cloud fully covers the camera-visible region, resulting in degraded image quality in areas without LiDAR measurements. More recent methods, including StreetGS~\cite{streetgs}, Omnire~\cite{omnire}, and~\cite{s3gs, letsgo}, attempt to integrate LiDAR data directly into the training pipeline by projecting the LiDAR depth point cloud onto the camera image plane to create a pseudo-depth image for supervision. However, this approach has limitations. First, the pseudo-depth image assumes depth measurements originate from the camera view, while in autonomous driving setups, LiDAR sensors are typically offset from the cameras. The side effect is shown in Figure~\ref{fig:lidar depth image loss}. Second, this method fails to leverage LiDAR points outside the camera’s field of view. To fully leverage LiDAR supervision in GS training, we propose a differentiable LiDAR model for GS that projects 3D Gaussians onto LiDAR range image frames. 

Concurrent works such as LiDAR-GS~\cite{lidargs} and GS-LiDAR~\cite{gslidar} propose LiDAR modeling focused exclusively on LiDAR data synthesis without image data synthesis capability, while \algname supports both image and LiDAR synthesis. SplatAD~\cite{splatad} is the work most similar to ours, modeling LiDAR and supporting both image and LiDAR synthesis. However, it focuses solely on urban scenes with close-range actors, whereas our method is validated on challenging highway scenes and supports long-range actor reconstruction. 


\section{Method}
\label{sec:method}
Figure~\ref{fig:system_diagram} illustrates an overview of the method. Section~\ref{sec:gaussian scene graph} presents the 3D Gaussian scene representation for dynamic scenes, including the LiDAR visibility for LiDAR rendering. Section~\ref{sec:camera modeling} discusses the camera modeling convention. Finally, Section~\ref{sec:lidar modeling} describes the proposed LiDAR modeling framework with four key components.

\subsection{3D Gaussian Scene Representation}
\label{sec:gaussian scene graph}
The scene is modeled as a set of 3D Gaussians, $\mathbf{G}$, also referred to as the splat model. $\mathbf{G}$ comprises of $N$ Gaussians, each with mean $\mu$, rotation quaternion $q$, scaling vector $S$, opacity $\alpha$, spherical harmonic (SH) coefficients $sh$, and LiDAR visibility rate $\gamma$: 
\begin{equation}
    \mathbf{G} = \{G_i:( \mu_i, q_i, S_i, \alpha_i, {sh}_i, \gamma_i)\ | i=1,...,N \}.
\end{equation}
Each Gaussian's covariance is parameterized by $\Sigma = RSS^TR^T$, where $S \in \mathbb{R}^3$ is a 3D scale vector with square roots of $\Sigma$'s eigenvalues and $R \in \text{SO}(3)$ is the rotation matrix computed from quaternion $q$.

\subsubsection{LiDAR Visibility}
\label{sec:liar_vis}
In addition to opacity $\alpha$ denoting the opaque state in an image, a LiDAR visibility rate $\gamma$ is introduced to handle the fundamental differences in how LiDAR and camera sensors perceive the environment. For example, LiDAR reflects off of surfaces based on their material and geometry. Low-reflective or semi-transparent surfaces can be missing in LiDAR measurements. Moreover, LiDAR has a limited sensing range compared to cameras. For example, the VLS-128 LiDAR can perceive only 5\% of targets $>$ 180m, whereas cameras can capture many more distant objects. Therefore, camera and LiDAR visibility are decoupled for each Gaussian via $\alpha_i^{L} = \alpha_i \gamma_i$ to better handle the challenging open highway scenes with many targets at far distances.


\subsubsection{Gaussian Scene Graph}
Dynamic driving scenes are reconstructed by separating the background and actors' splat models following the NSG~\cite{NSG} used in~\cite{drivinggs, streetgs, autosplat}. The time sequence of actor transformations, obtained from human-labeled bounding boxes, is then used to combine the background and actor splat models with means $\mu_{ij} = R_j \mu^o_{ij} + T_j$ and covariances $\Sigma_{ij} = R_j \Sigma^o_{ij} R_j^T$, 
where $\mu^o_{ij}$, $\Sigma^o_{ij}$ are the $i$th Gaussian mean and covariance in the $j$th object model, and $R_j$, $T_j$ are object rotation and translation. 

\subsection{Camera Modeling for Gaussian Splatting}
\label{sec:camera modeling}
\subsubsection{Color Rendering} To render an image from a camera view from the Gaussian primitive, we need to project 3D Gaussians into a 2D image plane as follows:
\begin{equation}
\mu^{I}_i = K W \mu_i
\end{equation}
\begin{equation}
\Sigma^{I}_i = J W \Sigma_i W^T J^T
\end{equation}
where $W$ is the camera pose with respect to the world frame, $K$ is the camera intrinsic matrix, and $J$ is the Jacobian of the projective transformation. The pixel color is computed as:
\begin{equation}
    \hat{\mathbf{C}}(p) = \sum_{i \in N} c_i \alpha'_i \prod_{j=1}^{i-1}(1-\alpha'_j)
\end{equation}
where $p$ is the pixel in an image, and $c_i$ is the color of a Gaussian obtained using $sh_i$ and view direction $v_{view}$. The opacity of a Gaussian at pixel $p$ is $\alpha'$, defined as: 
\begin{equation}
\alpha'_i = \alpha_i \exp\left({-\frac{1}{2}(p-\mu^{I}_i)^T {\Sigma^{I}}_i ^{-1}(p-\mu^{I}_i)}\right)
\end{equation}
where $\mu^{I}_i$ and $\Sigma^{I}_i$ are the Gaussian center and covariance in projected 2D image space. The rasterization process is denoted as $\bm{\pi}$, and the rendered color image is $\hat{\mathbf{C}} = \bm{\pi}(c, \alpha, \mu^{I}, \Sigma^{I})$.

\subsubsection{Depth Rendering} Recent studies project LiDAR point cloud into the image plane to generate pseudo-depth images $D$ for GS training~\cite{streetgs, s3gs, letsgo}. The rendered depth image is $\hat{D} = \bm{\pi}(d, \alpha, \mu^{I}, \Sigma^{I})$
, 
where $d$ is the depth of the Gaussian center. 
The depth loss is computed as $ \mathcal{L}_{depth} = \lVert \mathbf{D} - \hat{\mathbf{D}} \rVert_1\ $.

\subsection{LiDAR Modeling for Gaussian Splatting}
\label{sec:lidar modeling}
\subsubsection{Range Image Rendering}
To render LiDAR range images from 3D Gaussians, we first convert 3D Gaussians from the original Cartesian coordinates to spherical coordinates.
The mean $\mu=(x, y, z)$ is converted as: 
\begin{equation}
\mu^{sph} = 
\begin{bmatrix}
r  \\ \theta \\ \phi
\end{bmatrix}
=\begin{bmatrix}
\sqrt{x^2 + y^2 + z^2}  \\ 
\arctan2(y, x) \\ 
\arcsin\left(\frac{z}{r}\right) 
\end{bmatrix}
\label{eq:cart2sph}
\end{equation}
and the 3D covariance matrix $\Sigma$ is converted as: 
\begin{equation}
\Sigma^{sph} = J \Sigma J^T
\end{equation}
where J is the Jacobian of Eq.~\ref{eq:cart2sph}:
\begin{equation}
J=
\begin{bmatrix}
\frac{x}{r} & \frac{y}{r} & \frac{z}{r} \\
\frac{-y}{x^2 + y^2} & \frac{x}{x^2 + y^2} & 0 \\
\frac{-x z}{r^2\sqrt{r^2 - z^2}} & \frac{-y z}{r^2\sqrt{r^2 - z^2}} & \frac{\sqrt{r^2 - z^2}}{r^2} \\
\end{bmatrix}
\end{equation}

Next, we project 3D Gaussians into the LiDAR range image frame for rasterization. 
The LiDAR range image $\mathbf{R} \in \mathbb{R}^{\mathcal{M} \times \mathcal{W}}$ is collected from $\mathcal{M}$-beam LiDAR with inclination angles $\Phi = \{\phi_1, \phi_2, \ldots, \phi_\mathcal{M}\}$ and azimuth resolution $\mathcal{W}$. The Gaussian mean in the range image is:
\begin{equation}
\mu^{L} = 
\begin{bmatrix}
v \\
u
\end{bmatrix}
=
\begin{bmatrix}
\arg \min\limits_{i} |\phi - \phi_i| \\
\frac{\theta \mathcal{W}}{2\pi}
\end{bmatrix}
\end{equation}
The Gaussian covariance in the LiDAR range image is:
\begin{equation}
\Sigma^{L} = A \Sigma^{sph} A^T
\end{equation}
\begin{equation}
A = \begin{bmatrix}
\frac{\text{1}}{\Delta \theta} & 0 \\
 0 & \frac{\mathcal{W}}{2\pi} 
\end{bmatrix}
\end{equation}
Given the non-uniform inclination beam angle distribution, the Jacobian $A$ is dependent on the elevation resolution $\Delta \theta_v$. 

Finally, the LiDAR range image is rendered as:
\begin{equation}
\hat {\mathbf{R}} = \bm{\pi}(d, \alpha^{L}, \mu^{L}, \Sigma^{L}),
\end{equation}
where $\alpha^{L}$ is LiDAR visibility introduced in Sec.~\ref{sec:liar_vis}.

\subsubsection{2D Gaussian Scale Compensation}
In the original GS image rasterization model, a small Gaussian far from the camera image plane becomes invisible in the projected pixel due to numerical instability. However, since LiDAR emits laser beams to measure the distance of objects, a Gaussian should remain fully visible if its center is sufficiently close to the ray, regardless of its scale. To address Gaussians that are lost during projection, we rescale the 2D Gaussian when it is near the ray and has a scale below the visible threshold. Eigen decomposition is applied to Gaussians to get their 2D scales after projection
$
   \Sigma^L_i = R^L_i S^L_i {R^L_i}^{-1}, \text{where } S^L_i = \operatorname{diag}(s_{i1}^2, s_{i2}^2)
$.

Visible scale is dependent on the distance of the Gaussian:
\begin{equation}
   s_{vis}^i = d_i \tan\left( \dfrac{2\pi}{\mathcal{W}} \right)
\end{equation}
We adjust the Gaussian scale to be invisible if three standard deviations fall below the visible scale $s_{vis}$. 
\begin{equation}
   \tilde{s}_{ij} = 
   \begin{cases}
   \dfrac{s_{vis}^i}{3}, & \text{if } s_{ij} < \dfrac{s_{vis}^i}{3} \\[10pt]
   s_{ij}, & \text{otherwise}
   \end{cases}
\end{equation}
Finally, the 2D covariance matrix is constructed using the updated scale $\tilde{\Sigma}^{L}_i = R^L_i \operatorname{diag}(\tilde{s}_{i1}, \tilde{s}_{i2}) {R^L}^{-1}$.


\begin{figure}[t!]
    \centering
    \vspace{0.05in}
    \includegraphics[width=0.95\linewidth]{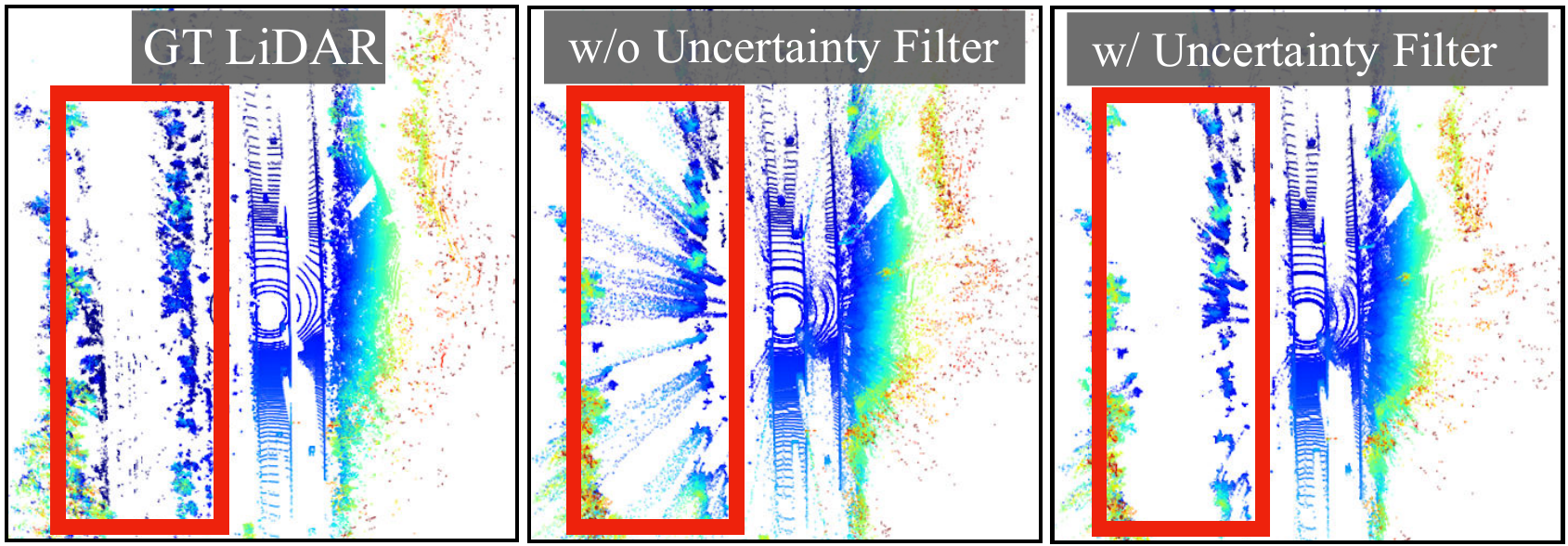}
        \vspace{-0.1in}
        \caption{\footnotesize Depth uncertainty filter removes floating artifacts on object edges from the rendered point cloud.}
        \label{fig:depth uncertainty point cloud}
\end{figure}

\begin{figure}[t!]
    \centering
    \includegraphics[width=0.90\linewidth]{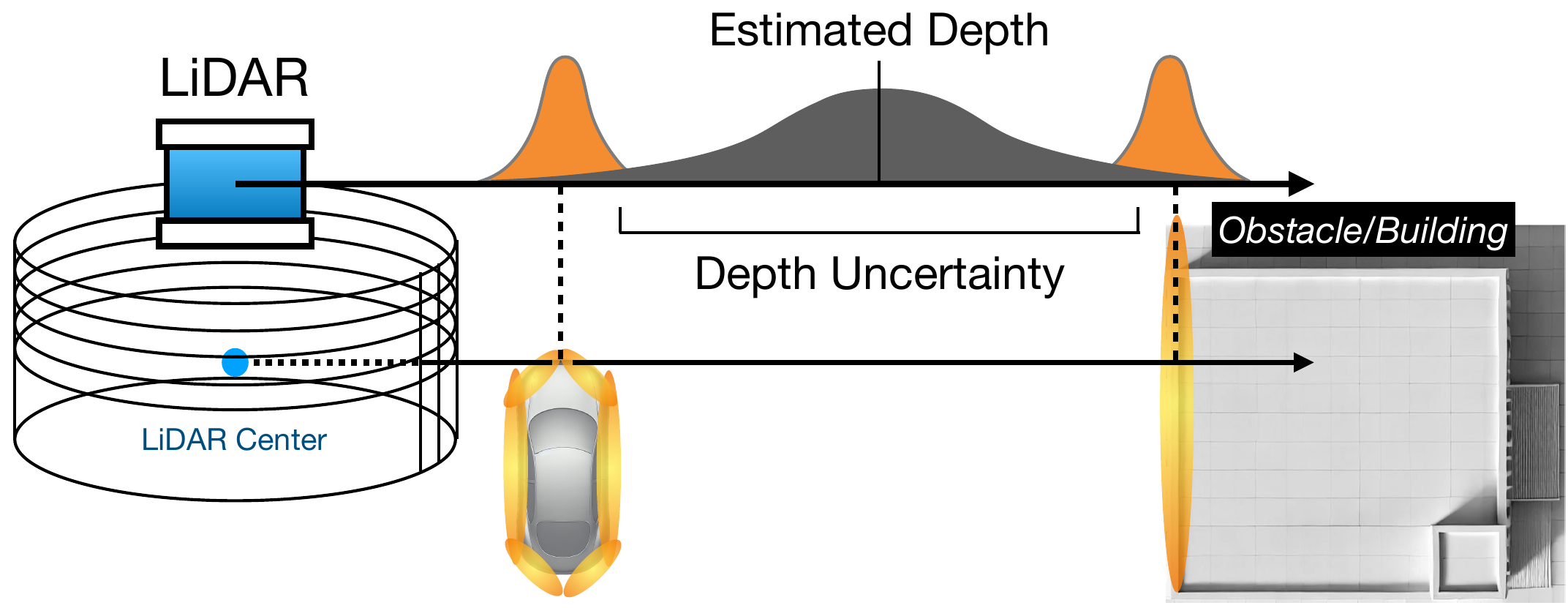}
        \caption{\footnotesize LiDAR depth rendering can lead to incorrect depth estimates at object edges, causing floating artifacts between adjacent objects. To address this, depth uncertainty is used to filter out the artifact on object edges, as shown in Figure~\ref{fig:depth uncertainty point cloud}.
        }
        \label{fig:depth uncertainty}
\end{figure}

\subsubsection{Depth Uncertainty Rendering}
We note that the point cloud rendered using GS can include floating artifacts on object edges (Figure~\ref{fig:depth uncertainty point cloud}). This is mainly because depth discontinuities are hard to represent using the inherently continuous Gaussian distribution. To render a crisp and clean point cloud with clear object edges, we introduce depth uncertainty rendering to identify pixels that cause floating noise in the rendered image, as shown in Figure~\ref{fig:depth uncertainty}. An uncertainty threshold is then applied to filter out floating points. Depth uncertainty is calculated as the incremental depth variance along the CUDA rasterization. 
The total accumulated alpha $\Lambda_i$ along a ray at the $i$-th Gaussian at a pixel is:
\begin{equation}
\Lambda_i = \Lambda_{i-1} + {\alpha^{L}}'_i,
\end{equation}
where $i$ is depth-sorted index, $\Lambda$ is initialized with $\Lambda_0=0$, and ${\alpha^{L}_n}'$ is the opacity contributed by a Gaussian at a pixel. 

Depth uncertainty $\Gamma_i$ can then be computed incrementally with Welford's online algorithm:
\begin{equation}
m_i = m_{i-1} + \frac{{\alpha^{L}}'_i}{\Lambda_i} \left( d_i - m_{i-1} \right)
\end{equation}
\begin{equation}
\Gamma_i = \Gamma_{i-1} + {\alpha^{L}}'_i \cdot \frac{\Lambda_{i-1}}{\Lambda_i} (d_i - m_{i-1})(d_i - m_{i-1})^T
\end{equation}
where $m_i$ is the running mean. An uncertainty threshold $\Gamma_{th}$ is then applied to filter out pixels with large depth uncertainty. Figure~\ref{fig:depth uncertainty point cloud}  demonstrates the rendered point cloud with and without the uncertainty filter.

\begin{figure}[t!]
    \centering
    \vspace{0.05in}
    \includegraphics[width=0.95\linewidth]{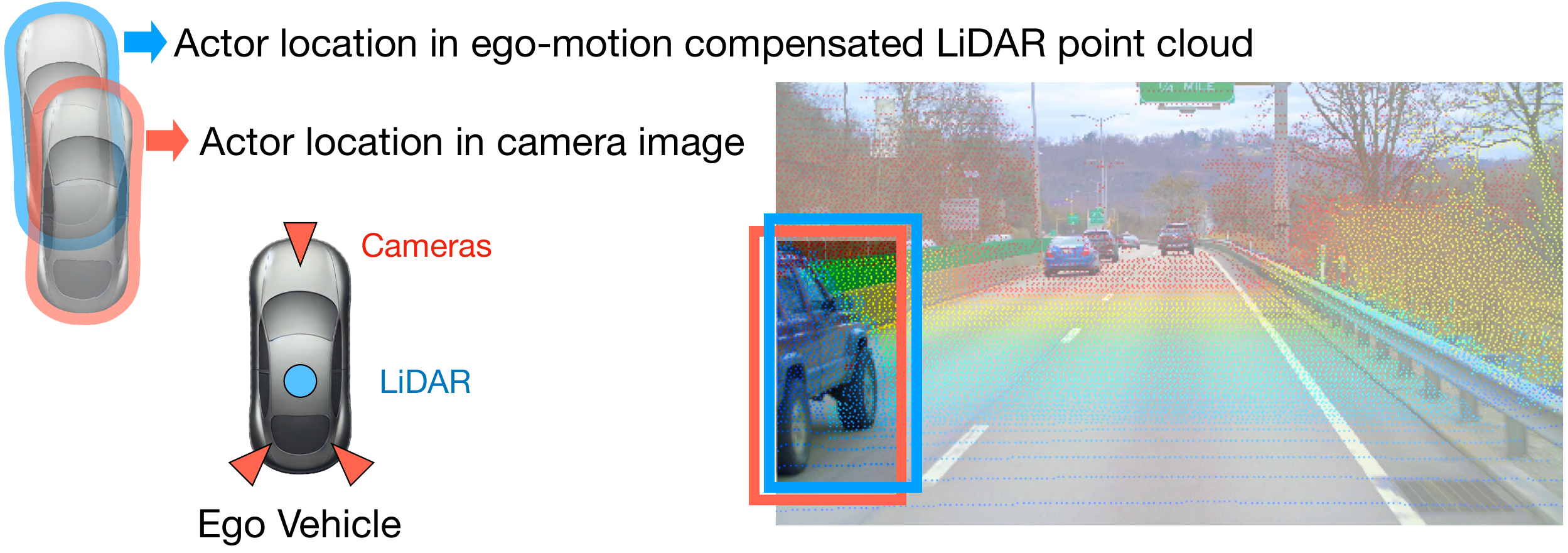}
        \vspace{-0.1in}
        \caption{\footnotesize Camera-LiDAR actor pose misalignment for high-speed actors.}
        \vspace{-0.2in}
        \label{fig:lidar-camera pose optimization}
\end{figure}

\begin{figure*}[t!]
    \centering
    \vspace{0.05in}
    \includegraphics[width=0.95\linewidth]{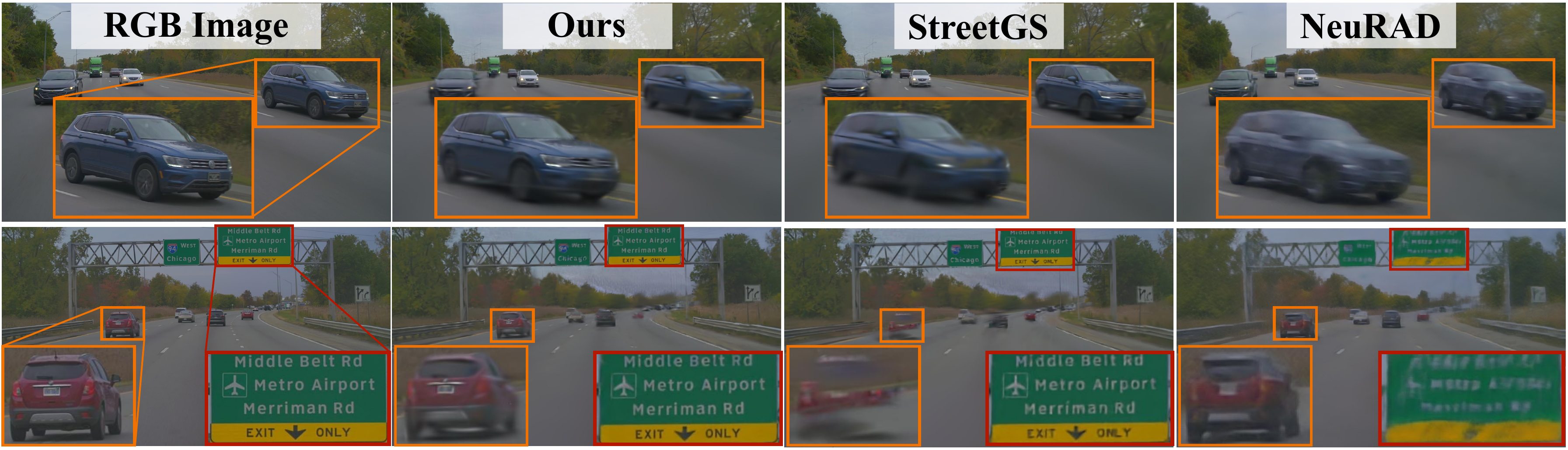}
        \vspace{-0.2cm}
        \caption{\footnotesize Qualitative comparison of image rendering. \algname delivers the best overall image quality, successfully capturing details on moving actors and a distant traffic sign, whereas the results from StreetGS and NeuRAD appear blurred.}
        \label{fig:render}
\end{figure*}

\begin{figure*}[t!]
    \centering
    \includegraphics[width=0.95\linewidth]{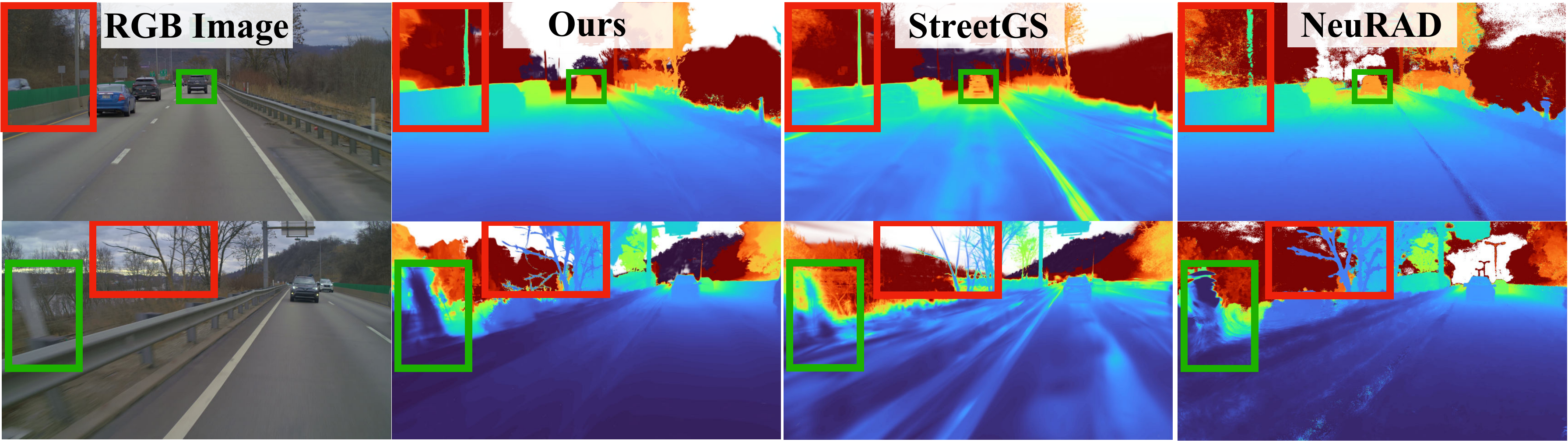}
        \vspace{-0.2cm}
        \caption{\footnotesize Qualitative comparison of depth rendering. Results show cleaner geometry over StreetGS due to the added LiDAR supervision. NeuRAD depth also has overall correct geometry, but is noisier. Using a 500m depth threshold to classify the sky (white) showed that NeuRAD introduces noisy depth measurements on the sky and fails to model the distant mountain.}
        \label{fig:rendered depth}
        \vspace{-0.25in}
\end{figure*}

\subsubsection{Camera-LiDAR Decoupled Pose Optimization}
Our method heavily relies on accurately labeled poses for actor reconstruction. However, human annotations can be imperfect, and the camera and LiDAR measurements can also be misaligned for fast-moving actors due to the temporal observation difference, as shown in Figure~\ref{fig:lidar-camera pose optimization}. Unlike~\cite{neurad, streetgs}, which optimize a unified pose, we propose a decoupled pose optimization to account for pose misalignment between sensors. Thus, the GS model for LiDAR and camera rendering is constructed separately using different poses $T^{j}_{lidar}$ and $T^{j}_{camera}$.

\subsection{Training Losses}
The total optimization objective is as follows:
\begin{equation}
\mathcal{L} = \mathcal{L}_{c} + \lambda_1\mathcal{L}_{lidar} + \lambda_2\mathcal{L}^{L}_{opa} + \lambda_3\mathcal{L}^s_{reg} + \lambda_4\mathcal{L}^{c}_{opa}
\end{equation}
\begin{equation}
\mathcal{L}^c_{opa} = \mathcal{L}_{sky} + \mathcal{L}_{fg} + \mathcal{L}^{obj}_{reg}
\end{equation}

\begin{table}[b]
\centering
\setlength{\tabcolsep}{2pt}
\renewcommand{\arraystretch}{1.3}

\resizebox{0.95\linewidth}{!}{%
\begin{tabular}{clcccclcc}
\hline
\multicolumn{9}{c}{Our Dataset (Highway Scenes)}                                                                                                                    \\ \hline
\multirow{2}{*}{Method}       &  & \multicolumn{4}{c}{Image}                                                                          &  & \multicolumn{2}{c}{LiDAR}           \\
                              &  & \multicolumn{1}{l}{PSNR$^\dagger$$\uparrow$} & PSNR$\uparrow$ & LPIPS$\downarrow$ & SSIM$\uparrow$ &  & Mean$\downarrow$ & Med.$\downarrow$ \\ \cline{1-1} \cline{3-6} \cline{8-9} 
Instant-NGP~\cite{instantngp} &  & 18.73                                        & 25.94          & 0.37              & 0.856          &  & -                & -                \\
3DGS~\cite{3dgs}              &  & 17.60                                        & 27.36          & 0.32              & 0.918          &  & -                & -                \\
NeuRAD~\cite{neurad}          &  & 19.87                                        & \second 30.38  & \second 0.28      & 0.911          &  & \third 2.30     & \best 0.38       \\
StreetGS~\cite{streetgs}      &  & \second 24.56                                & 29.44          & \third 0.32              & \third 0.921   &  & 8.70*            & 4.73*            \\
SplatAD~\cite{splatad}                       &  & \third 24.71                                & \third 29.89   & 0.34              & \second 0.925  &  & \second 2.23               & \third 0.44               \\
\algname (Ours)               &  & \best 26.43                                  & \best 30.69    & \best 0.27        & \best 0.927    &  & \best 1.66       & \second 0.43     \\ \hline
\end{tabular}
}
\vspace{0.05in}
{ \fontsize{6.5pt}{7.5pt}\selectfont ~\\ *Rendering LiDAR with our proposed LiDAR rendering.}
\vspace{-0.05in}
\caption{\footnotesize Image and LiDAR novel view rendering on self-collected dataset. PSNR$^\dagger$ denotes the PSNR of moving objects. Best results in \best{red}, second best in \second{orange}, and third in \third{yellow}.}
\label{tab:render_avg_splatad}
\end{table}

\begin{table}[!b]
\centering
\vspace{-0.1in}
\setlength{\tabcolsep}{2pt}
\renewcommand{\arraystretch}{1.3}
\resizebox{0.9\linewidth}{!}{%
\begin{tabular}{clcccclcc}
\hline
\multicolumn{9}{c}{Waymo Open Dataset (Urban Scenes)}                                                                                                      \\ \hline
Method                        &  & \multicolumn{4}{c}{Image}                                                      &  & \multicolumn{2}{c}{LiDAR}           \\
                              &  & PSNR$^\dagger$$\uparrow$ & PSNR$\uparrow$ & LPIPS$\downarrow$ & SSIM$\uparrow$ &  & Mean$\downarrow$ & Med.$\downarrow$ \\ \cline{1-1} \cline{3-6} \cline{8-9} 
Instant-NGP~\cite{instantngp} &  & 25.09                    & 28.70          & 0.19              & 0.887          &  & -                & -                \\
3DGS~\cite{3dgs}              &  & 25.05                    & 28.86          & \best0.13         & 0.937          &  & -                & -                \\
DeformGS~\cite{deformgs}      &  & 21.42                    & 28.86          & 0.17              & 0.876          &  & -                & -                \\
PVG~\cite{pvg}                &  & 25.24                    & 29.23          & 0.19              & 0.863          &  & -                & -                \\
Omnire~\cite{omnire}          &  & 26.30                    & 29.22          & \second 0.15      & 0.866          &  & -                & -                \\
NeuRAD~\cite{neurad}          &  & 28.83                    & 29.26          & 0.20              & 0.844          &  & \second 0.69     & \second 0.07     \\
StreetGS~\cite{streetgs}      &  & \second 29.02            & \second 33.24  & \second 0.15      & \second 0.943  &  & 5.00*            & 2.08*            \\
LiHi-GS (Ours)                &  & \best 29.53              & \best 33.81    & \best 0.13        & \best 0.945    &  & \best 0.37       & \best 0.06       \\ \hline
\end{tabular}
}
\vspace{0.02in}
{ \fontsize{6.5pt}{7.5pt}\selectfont ~\\ *Rendering LiDAR with our proposed LiDAR rendering.}
\caption{ Image and LiDAR novel view rendering on the Waymo Open Dataset. PSNR$^\dagger$ denotes the PSNR of moving objects. 
Best results in red and second best in orange.}
\label{tab:waymo}
\vspace{-0.05in}
\end{table}

Color loss \(\mathcal{L}_{c}\) follows the color image loss from~\cite{3dgs}. LiDAR loss is calculated as the L1 difference between the target and rendered range images.
\(\mathcal{L}^L_{opa}\) encourages LiDAR visibility \(\alpha^{L} = 1\) for pixels with depth returns in the range image. This loss term is crucial considering LiDAR's 360° field of view, which includes regions outside camera coverage. \(\mathcal{L}^s_{reg}\) is a scale regularization term designed to make Gaussians more evenly shaped, preventing spike artifacts and providing more accurate depth rendering, inspired by~\cite{physgaussian}. \(\mathcal{L}^c_{opa}\) encompasses all opacity losses from the camera image. \(\mathcal{L}_{sky}\) ensures sky pixels have low opacity, while \(\mathcal{L}_{fg}\) guides foreground pixels to have high opacity. Additionally, \(\mathcal{L}^{obj}_{reg}\) is a regularization term used to improve decomposition effects, following~\cite{streetgs}.

\section{Experiments}
\label{sec:experiments}
\subsection{Datasets}
Existing research often relies on open-source datasets for evaluation that are dominated by texture-rich urban scenarios with dense sensor coverage. This limits their applicability to feature-sparse and view-sparse highway scenarios, a key use case for autonomous driving. To address this gap, we conduct experiments on a self-collected dataset with challenging highway environments. Our data collection vehicle is equipped with a VLS-128 LiDAR and three cameras facing front, back-left, and back-right, synchronized at 10 Hz. We gather highway data across three U.S. states (Michigan, Pennsylvania, and North Carolina). 
Four challenging segments, ranging from 16 to 25 seconds in length and featuring multiple moving objects, high ego speeds, small-scale far-field objects, and monotonous backgrounds, are used in our experiments. 
We sampled every 10th frame for evaluation, and used the remaining frames exclusively for training.

Additionally, to evaluate \algname on urban scenes and public datasets, we conduct experiments on the Waymo open dataset~\cite{waymo} on sequences used in~\cite{streetg_ns} and pandaSet~\cite{pandaset} and nuScenes~\cite{nuscenes} datasets on sequences used in~\cite{splatad}.

\subsection{Baselines}
\algname is compared with multiple baselines on both our dataset and the Waymo dataset. On our dataset, we compare against: 
Instant-NGP~\cite{instantngp}, an image-only NeRF-based method;
NeuRAD~\cite{neurad}, a NeRF-based method for driving scenes that supports LiDAR supervision and rendering;
and StreetGS~\cite{streetgs}, a GS-based method designed for autonomous driving. 
For 3DGS~\cite{3dgs}, we use a re-implementation from splatfacto~\cite{gsplat}.
We also deploy the official implementation of concurrent work, SplatAD~\cite{splatad}, on our highway dataset.

On the Waymo dataset, we perform benchmarking experiments with Omnire~\cite{omnire}, PVG~\cite{pvg}, and DeformGS~\cite{deformgs} using~\cite{omnire} implementation, which supports the Waymo data format.
We also compare with GS-LiDAR~\cite{gslidar}, on LiDAR-only novel view rendering on Waymo sequences used in \cite{gslidar}. 

We additionally provide comparisons on the PandaSet and nuScenes datasets with baselines reported in the SplatAD~\cite{splatad}, which are included on the \href{https://umautobots.github.io/lihi_gs}{project page}.

\begin{figure}[b!]
    \centering
    \includegraphics[width=0.99\linewidth]{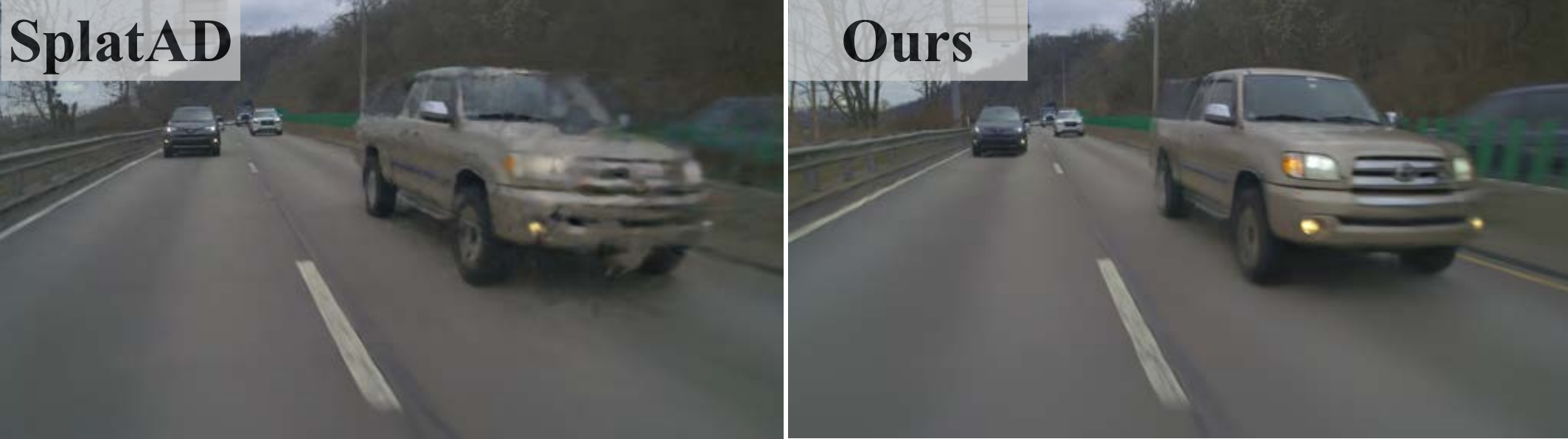}
    \vspace{-5mm}
    \caption{\footnotesize Qualitative comparison of image rendering on our highway dataset. \algname produces sharper and cleaner reconstructions of moving vehicles compared to the blurry results of SplatAD, demonstrating the effectiveness of decoupled camera–LiDAR pose optimization for moving actor reconstruction.}
    \vspace{-1mm}
    \label{fig:render_vs_splatad}
\end{figure}

\begin{figure}[b!]
    \centering
    \includegraphics[width=1.\linewidth]{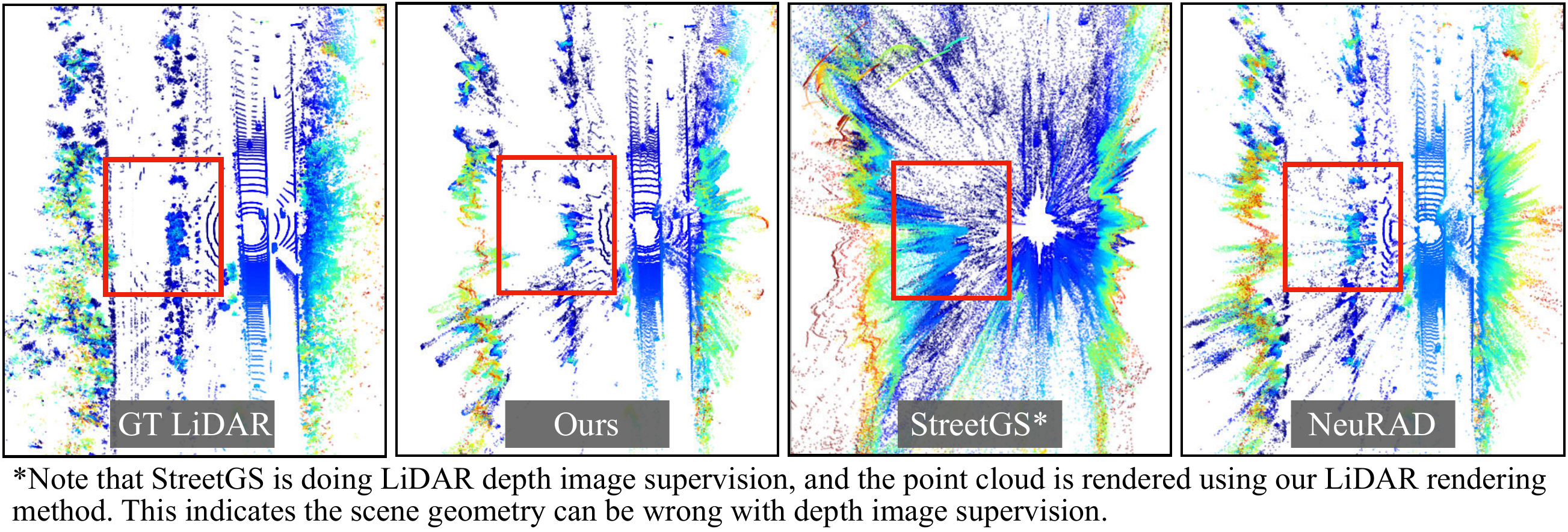}
        \vspace{-0.2in}
        \caption{\footnotesize Point cloud synthesized using different methods. Our method provides a clean point cloud with minimal noise on object edges.}
        \label{fig:rendered point cloud}
        \vspace{-0.1in}
\end{figure}

\begin{figure}[b!]
    \centering
    \includegraphics[width=1.0\linewidth]{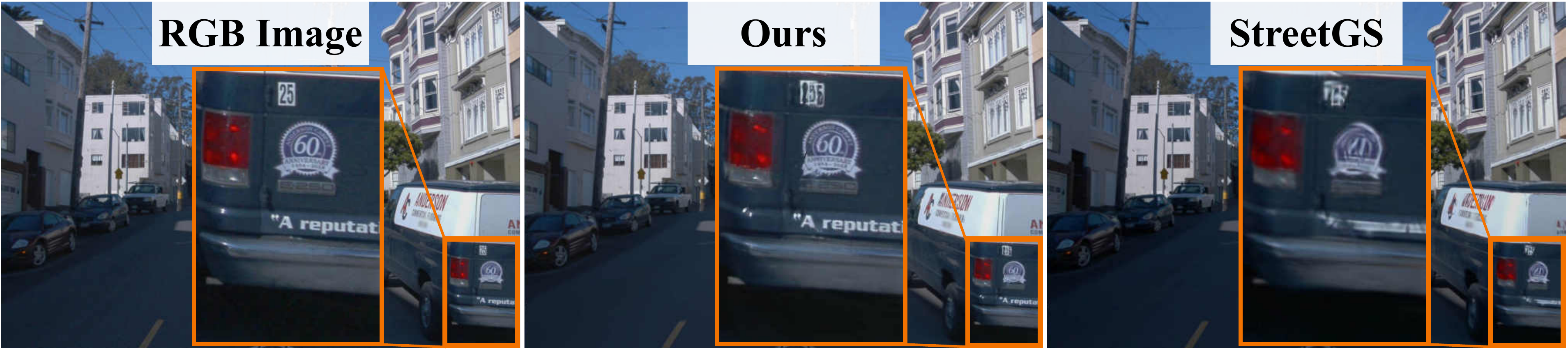}
    \vspace{-3mm}
    \caption{\footnotesize Qualitative comparison of novel view rendering on the Waymo Dataset. \algname captures clear details on the vehicle.}
    \vspace{-2mm}
    \label{fig:waymo}
\end{figure}

\begin{figure*}[ht!]
    \centering
    \vspace{0.05in}
    \includegraphics[width=0.9\linewidth]{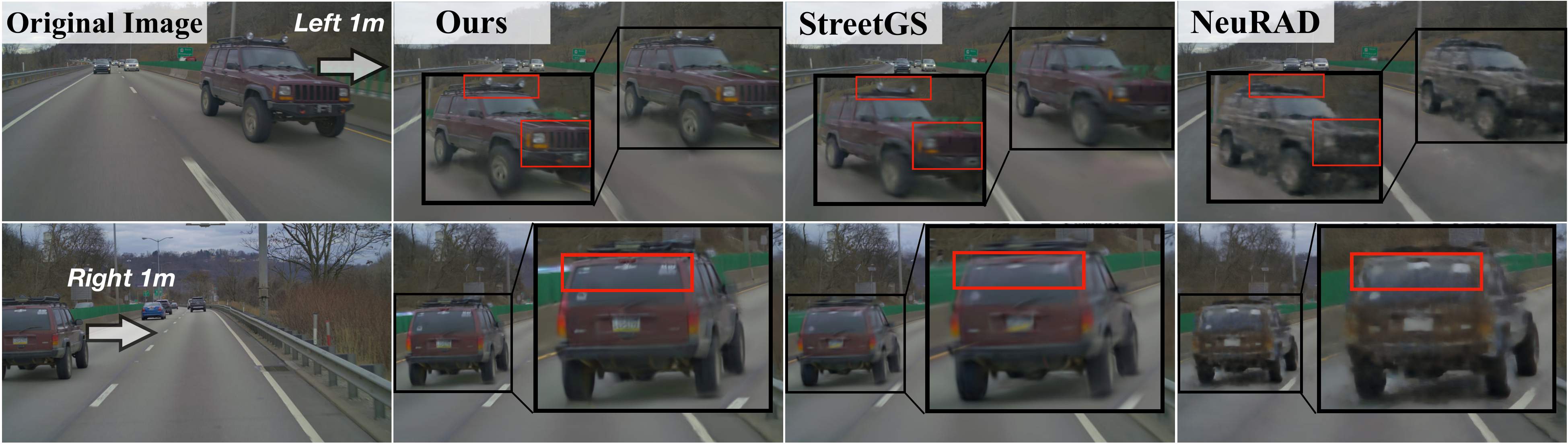}
        \vspace{-0.3cm}
        \caption{\footnotesize Scene editing with lateral actor shift. \algname has the best overall image quality, given the accurate geometry learned from LiDAR supervision.}
        \label{fig:actor shift}
\end{figure*}

\subsection{Novel View Image and LiDAR Synthesis}
Table~\ref{tab:render_avg_splatad} shows the quantitative comparison of our \algname with SOTA baselines on our highway data for novel view rendering quality. \algname outperforms all baselines on all image metrics including moving object reconstruction, highlighting the effectiveness of the proposed LiDAR supervision. \algname also surpasses SplatAD, demonstrating the importance of the proposed decoupled pose optimization and LiDAR visibility components, which are specifically designed for highway scenes. 
Qualitative comparison of rendered image and depth is shown in Figures~\ref{fig:render} and ~\ref{fig:rendered depth}. Comparison with SplatAD is shown in Figure~\ref{fig:render_vs_splatad}. \algname captures finer details in the rendered image and accurately renders the scene geometry in rendered depth.
In terms of LiDAR rendering performance, \algname consistently has the lowest L1 mean error due to the decoupled pose optimization and uncertainty filter that is lacking in NeuRAD, as shown in Figure~\ref{fig:rendered point cloud}. 
The actor/ego shift experiments (Table~\ref{tab:scene edit}) demonstrate that \algname yields better geometry reconstruction. 


Table~\ref{tab:waymo} shows the quantitative results on the public Waymo Open Dataset~\cite{waymo}. We observed a similar trend of improvement as experiments conducted on our self-collected dataset. \algname outperforms existing works on both image and LiDAR novel view synthesis. 
The overall improvement of \algname on Waymo urban data is smaller than in highway scenes, suggesting LiDAR supervision is more critical in challenging highway scenarios. However, the qualitative results in Figure~\ref{fig:waymo} show that \algname still captures more details with LiDAR supervision on urban scenes. The image and LiDAR rendering achieve 72 and 5 FPS on a NVIDIA A100 GPU.

We report additional results on the nuScenes~\cite{nuscenes} and PandaSet~\cite{pandaset} datasets on our \href{https://umautobots.github.io/lihi_gs}{project
 page}, demonstrating LiHi-GS’s strong generalization across public datasets and outperforming SplatAD, thanks to our camera–LiDAR decoupled pose optimization and LiDAR visibility rate.

Table~\ref{tab:gslidar} compares LiDAR-only \algname with concurrent GS-LiDAR~\cite{gslidar} on the Waymo Dataset. \algname provides better LiDAR range image synthesis and offers a competitive median error while providing extra camera rendering.

\subsection{Scene Editing With Ego and Actor Shifting}
\begin{figure}[t!]
    \centering
    \includegraphics[width=0.85\linewidth]{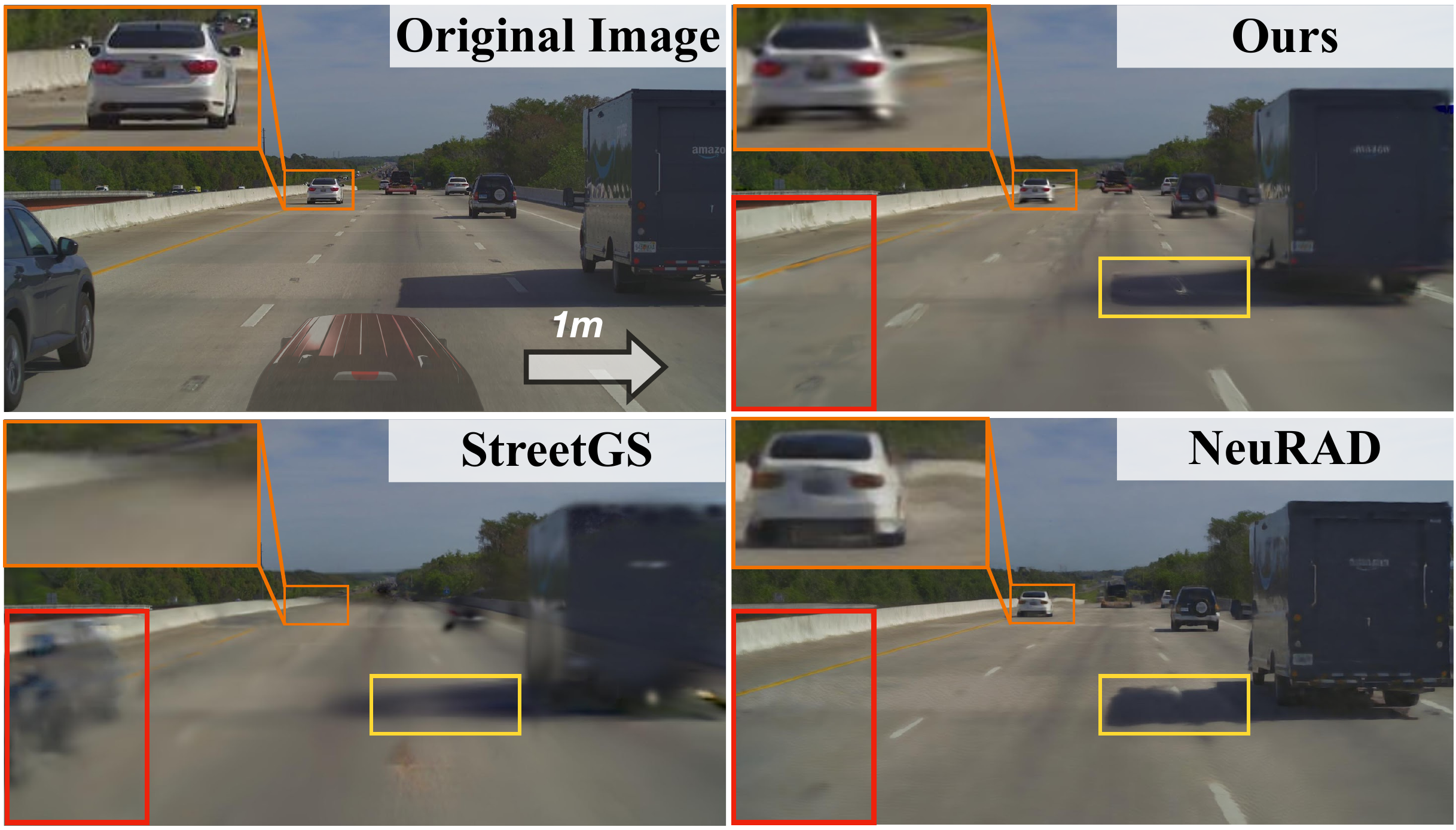}
        \vspace{-0.1in}
        \caption{\footnotesize Scene editing with lateral ego-vehicle shift. \algname reconstructs vehicles with correct color, better shadow shape, and fewer ghost artifacts.}
        \label{fig:ego shift}
\end{figure}

We further evaluate the rendering quality with ego-vehicle and surrounding actors shifting. We apply lateral shifting for both ego and actors from 1 meter to 3 meters. Since the ground truth after scene editing does not exist, we follow~\cite{neurad} to report the FID score to analyze the data synthesis quality. Table~\ref{tab:scene edit} shows that \algname generates the most realistic image from new viewpoints that are off from the original trajectory and learns a better representation of actors. The qualitative results of actor and ego shifting are shown in Figures~\ref{fig:actor shift} and~\ref{fig:ego shift}.



\subsection{Ablations}
Table~\ref{tab:ablations} presents the quantitative results when removing the different proposed designs from our pipeline. 
Removing LiDAR opacity loss $\mathcal{L}^{L}_{opa}$ leads to the highest LiDAR L1 mean error. 
Scale regularization $\mathcal{L}^{s}_{reg}$ also has a great impact on both image and LiDAR quality. Rendered depth without scale regularization is inaccurate and can degrade the image quality during LiDAR supervision. Disabling decoupled pose optimization leads to camera-LiDAR pose misalignment and worse PSNR rendering quality. 
The model without 2D scale compensation crashed midway through training due to memory overhead. Without scale compensation, LiDAR supervision inflates the size of distant Gaussians to match the projected LiDAR pixels, leading to unrealistic geometry and excessive memory usage. In the end, LiDAR visibility significantly improves both image and LiDAR rendering quality. This reflects the fundamental differences between the two sensors. 

We also compare the impact of the proposed loss with depth image loss used in prior works. The results show that depth image loss does not enhance image rendering quality and can sometimes degrade it.
In contrast, our LiDAR supervision improves both image rendering quality and LiDAR accuracy. 

\begin{table}[!t]
\centering
\vspace{-0.1in}
\setlength{\tabcolsep}{6pt}
\renewcommand{\arraystretch}{1.2}
\resizebox{0.8\linewidth}{!}{%
\begin{tabular}{clcccc}
\hline
\multicolumn{6}{c}{LiDAR-Only Evaluation}                                                                                         \\ \hline
\multirow{2}{*}{Method} &  & \multicolumn{4}{c}{LiDAR Range Image}                                                                                      \\
                        &  & PSNR$\uparrow$ & \multicolumn{1}{l}{LPIPS$\downarrow$} & \multicolumn{1}{l}{SSIM$\uparrow$} & Med.$\downarrow$ \\ \cline{1-1} \cline{3-6} 
GS-LiDAR~\cite{gslidar}                &  & 22.25          & 0.07                                  & 0.839                              & \best 0.020            \\
LiHi-GS (Ours)          &  & \best 23.00          & \best 0.05                                  & \best 0.894                              & 0.023            \\ \hline
\end{tabular}
}
\caption{ Comparing LiDAR novel view rendering with GS-LiDAR on Waymo sequences used in~\cite{gslidar} with LiDAR-only training.
}
\label{tab:gslidar}
\vspace{-0.2in}
\end{table}

\begin{table}[!t]
\centering
\setlength{\tabcolsep}{5pt}
\renewcommand{\arraystretch}{1.2}
\resizebox{0.85\linewidth}{!}{%
\begin{tabular}{cccclccc}
\hline
\textbf{} & \multicolumn{3}{c}{Ego Lateral Shift (FID)} & \textbf{} & \multicolumn{3}{c}{Actor Lateral Shift (FID)} \\ \cline{2-4} \cline{6-8} 
\textbf{} & 1m & 2m & 3m & \textbf{} & 1m & 2m & 3m \\ \hline
NeuRAD & 63.2 & 82.42 & 101.45 &  & 51.25 & 55.61 & 79.70 \\
StreetGS & 56.09 & 69.59 & 82.24 &  & 43.65 & 54.48 & 65.86 \\
\algname & \best 40.70 & \best 57.47 & \best 80.69 &  & \best 30.91 & \best 43.72 & \best 54.50 \\ \hline
\end{tabular}
}
\caption{\footnotesize FID ($\downarrow$) scores when shifting ego vehicle or actors pose.}
\label{tab:scene edit}
\vspace{-0.2in}
\end{table}

\begin{table}[!t]
\centering
\setlength{\tabcolsep}{6pt}
\renewcommand{\arraystretch}{1.2}
\newcolumntype{C}[1]{>{\centering\arraybackslash}p{#1}}
\resizebox{0.95\linewidth}{!}{%
\begin{tabular}{lcccc}
\hline
\multicolumn{1}{c}{\multirow{2}{*}{Module Ablations}}                       & \multicolumn{3}{c}{Image}                                          & LiDAR                \\
\multicolumn{1}{c}{}                                                & PSNR$\uparrow$       & IPIPS$\downarrow$    & SSIM$\uparrow$       & Mean$\downarrow$     \\ \hline
w/o $\mathcal{L}^L_{opa}$                                           & 30.54                & 0.229                & 0.927                & 2.94                 \\
w/o $\mathcal{L}^s_{reg}$                                           & 29.65                & 0.234                & 0.926                & 2.76                 \\
w/o Decoupled pose optimization                                     & 30.49                & 0.228                & 0.927                & \second 2.67         \\
w/o 2D scale compensation                                           & 30.22*               & 0.245*               & 0.923*               & 2.71*                \\
w/o LiDAR visibility                                                & \second 30.59        & \second 0.228        & \second 0.928        & \second 2.67         \\
\textbf{Proposed Full}                                              & \best 31.63          & \best 0.208          & \best 0.935          & \best 2.61           \\ \hline
\multicolumn{1}{c}{LiDAR Loss Ablations}                             & \multicolumn{1}{l}{} & \multicolumn{1}{l}{} & \multicolumn{1}{l}{} & \multicolumn{1}{l}{} \\ \hline
Only $\mathcal{L}_c$~\cite{drivinggs}                                    & 29.59                & 0.293                & 0.923                & 8.66                \\
$+ \mathcal{L}_{depth}$~\cite{streetgs, s3gs, letsgo} & \second 29.70        & \second 0.285        & \second 0.925        & \second 6.23         \\
 $+\ \text{Proposed }\mathcal{L}_{\text{lidar}}$
        & \best 31.63          & \best 0.208          & \best 0.935          & \best 2.61           \\ \hline
\end{tabular}
}
\vspace{-0.in}
\begin{flushleft}
\scriptsize{*Disabling 2D scale compensation caused an exponential increase in memory usage, leading the job to crash midway.}
\end{flushleft}
\vspace{-0.1in}
\caption{\footnotesize Ablation studies.}
\vspace{-0.4in}
\label{tab:ablations}
\end{table}

\section{Conclusions}
\label{sec:conclusion}
In summary, we introduce a differentiable GS LiDAR rendering model for challenging highway scenes. Our results demonstrate that \algname achieves SOTA performance in novel view synthesis. We highlight the importance of LiDAR supervision in learning accurate scene geometry, which significantly enhances image rendering quality, especially under actor or ego shifts. Unlike prior works that primarily focus on urban datasets, our approach is the first to evaluate on monotonous highway data, bridging the gap between existing research and real-world autonomous driving applications. 



{\fontsize{7.5}{9.3}\selectfont
\bibliographystyle{IEEEtran}
\bibliography{main}

@String(TOG= {ACM Trans. Graph.})

@String(TOG   = {ACM TOG})

@article{gslidar,
  title={GS-LiDAR: Generating Realistic LiDAR Point Clouds with Panoramic Gaussian Splatting},
  author={Jiang, Junzhe and Gu, Chun and Chen, Yurui and Zhang, Li},
  journal={arXiv preprint arXiv:2501.13971},
  year={2025}
}

@article{splatad,
  title={SplatAD: Real-Time Lidar and Camera Rendering with 3D Gaussian Splatting for Autonomous Driving},
  author={Hess, Georg and Lindstr{\"o}m, Carl and Fatemi, Maryam and Petersson, Christoffer and Svensson, Lennart},
  journal={arXiv preprint arXiv:2411.16816},
  year={2024}
}

@inproceedings{waymo,
  title={Waymo open dataset: Panoramic video panoptic segmentation},
  author={Mei, Jieru and Zhu, Alex Zihao and Yan, Xinchen and Yan, Hang and Qiao, Siyuan and Chen, Liang-Chieh and Kretzschmar, Henrik},
  booktitle={European Conference on Computer Vision},
  pages={53--72},
  year={2022},
  organization={Springer}
}

@article{omnire,
  title={Omnire: Omni urban scene reconstruction},
  author={Chen, Ziyu and Yang, Jiawei and Huang, Jiahui and de Lutio, Riccardo and Esturo, Janick Martinez and Ivanovic, Boris and Litany, Or and Gojcic, Zan and Fidler, Sanja and Pavone, Marco and others},
  journal={arXiv preprint arXiv:2408.16760},
  year={2024}
}

@inproceedings{carla,
  title={CARLA: An open urban driving simulator},
  author={Dosovitskiy, Alexey and Ros, German and Codevilla, Felipe and Lopez, Antonio and Koltun, Vladlen},
  booktitle={Conference on robot learning},
  pages={1--16},
  year={2017},
  organization={PMLR}
}

@misc{teslacrash,
  title        = {Tesla Autopilot feature was involved in 13 fatal crashes, US regulator says},
  url          = {https://www.theguardian.com/technology/2024/apr/26/tesla-autopilot-fatal-crash},
}

@misc{streetg_ns,
  author       = {Lightwheel AI},
  title        = {Street-Gaussians-ns},
  year         = {2024},
  url          = {https://github.com/LightwheelAI/street-gaussians-ns/tree/main},
}

@article{nerf,
  title={Nerf: Representing scenes as neural radiance fields for view synthesis},
  author={Mildenhall, Ben and Srinivasan, Pratul P and Tancik, Matthew and Barron, Jonathan T and Ramamoorthi, Ravi and Ng, Ren},
  journal={Communications of the ACM},
  volume={65},
  number={1},
  pages={99--106},
  year={2021},
  publisher={ACM New York, NY, USA}
}

@inproceedings{meganerf,
  title={Mega-nerf: Scalable construction of large-scale nerfs for virtual fly-throughs},
  author={Turki, Haithem and Ramanan, Deva and Satyanarayanan, Mahadev},
  booktitle={Proceedings of the IEEE/CVF Conference on Computer Vision and Pattern Recognition},
  pages={12922--12931},
  year={2022}
}

@inproceedings{blocknerf,
  title={Block-nerf: Scalable large scene neural view synthesis},
  author={Tancik, Matthew and Casser, Vincent and Yan, Xinchen and Pradhan, Sabeek and Mildenhall, Ben and Srinivasan, Pratul P and Barron, Jonathan T and Kretzschmar, Henrik},
  booktitle={Proceedings of the IEEE/CVF Conference on Computer Vision and Pattern Recognition},
  pages={8248--8258},
  year={2022}
}

@article{instantngp,
  title={Instant neural graphics primitives with a multiresolution hash encoding},
  author={M{\"u}ller, Thomas and Evans, Alex and Schied, Christoph and Keller, Alexander},
  journal={ACM transactions on graphics (TOG)},
  volume={41},
  number={4},
  pages={1--15},
  year={2022},
  publisher={ACM New York, NY, USA}
}

@inproceedings{NSG,
  title={Neural scene graphs for dynamic scenes},
  author={Ost, Julian and Mannan, Fahim and Thuerey, Nils and Knodt, Julian and Heide, Felix},
  booktitle={Proceedings of the IEEE/CVF Conference on Computer Vision and Pattern Recognition},
  pages={2856--2865},
  year={2021}
}

@inproceedings{MARS,
  title={Mars: An instance-aware, modular and realistic simulator for autonomous driving},
  author={Wu, Zirui and Liu, Tianyu and Luo, Liyi and Zhong, Zhide and Chen, Jianteng and Xiao, Hongmin and Hou, Chao and Lou, Haozhe and Chen, Yuantao and Yang, Runyi and others},
  booktitle={CAAI International Conference on Artificial Intelligence},
  pages={3--15},
  year={2023},
  organization={Springer}
}

@inproceedings{SUDS,
  title={Suds: Scalable urban dynamic scenes},
  author={Turki, Haithem and Zhang, Jason Y and Ferroni, Francesco and Ramanan, Deva},
  booktitle={Proceedings of the IEEE/CVF Conference on Computer Vision and Pattern Recognition},
  pages={12375--12385},
  year={2023}
}

@article{emernerf,
  title={Emernerf: Emergent spatial-temporal scene decomposition via self-supervision},
  author={Yang, Jiawei and Ivanovic, Boris and Litany, Or and Weng, Xinshuo and Kim, Seung Wook and Li, Boyi and Che, Tong and Xu, Danfei and Fidler, Sanja and Pavone, Marco and others},
  journal={arXiv preprint arXiv:2311.02077},
  year={2023}
}

@inproceedings{lidar-nerf,
  title={Lidar-nerf: Novel lidar view synthesis via neural radiance fields},
  author={Tao, Tang and Gao, Longfei and Wang, Guangrun and Lao, Yixing and Chen, Peng and Zhao, Hengshuang and Hao, Dayang and Liang, Xiaodan and Salzmann, Mathieu and Yu, Kaicheng},
  booktitle={Proceedings of the 32nd ACM International Conference on Multimedia},
  pages={390--398},
  year={2024}
}

@inproceedings{NFL,
  title={Neural lidar fields for novel view synthesis},
  author={Huang, Shengyu and Gojcic, Zan and Wang, Zian and Williams, Francis and Kasten, Yoni and Fidler, Sanja and Schindler, Konrad and Litany, Or},
  booktitle={Proceedings of the IEEE/CVF International Conference on Computer Vision},
  pages={18236--18246},
  year={2023}
}

@inproceedings{dynNFL,
  title={Dynamic LiDAR Re-simulation using Compositional Neural Fields},
  author={Wu, Hanfeng and Zuo, Xingxing and Leutenegger, Stefan and Litany, Or and Schindler, Konrad and Huang, Shengyu},
  booktitle={Proceedings of the IEEE/CVF Conference on Computer Vision and Pattern Recognition},
  pages={19988--19998},
  year={2024}
}

@inproceedings{lidar4d,
  title={LiDAR4D: Dynamic Neural Fields for Novel Space-time View LiDAR Synthesis},
  author={Zheng, Zehan and Lu, Fan and Xue, Weiyi and Chen, Guang and Jiang, Changjun},
  booktitle={Proceedings of the IEEE/CVF Conference on Computer Vision and Pattern Recognition},
  pages={5145--5154},
  year={2024}
}

@inproceedings{urf,
  title={Urban radiance fields},
  author={Rematas, Konstantinos and Liu, Andrew and Srinivasan, Pratul P and Barron, Jonathan T and Tagliasacchi, Andrea and Funkhouser, Thomas and Ferrari, Vittorio},
  booktitle={Proceedings of the IEEE/CVF Conference on Computer Vision and Pattern Recognition},
  pages={12932--12942},
  year={2022}
}

@article{cloner,
  title={Cloner: Camera-lidar fusion for occupancy grid-aided neural representations},
  author={Carlson, Alexandra and Ramanagopal, Manikandasriram S and Tseng, Nathan and Johnson-Roberson, Matthew and Vasudevan, Ram and Skinner, Katherine A},
  journal={IEEE Robotics and Automation Letters},
  volume={8},
  number={5},
  pages={2812--2819},
  year={2023},
  publisher={IEEE}
}

@article{loner,
  title={Loner: Lidar only neural representations for real-time slam},
  author={Isaacson, Seth and Kung, Pou-Chun and Ramanagopal, Mani and Vasudevan, Ram and Skinner, Katherine A},
  journal={IEEE Robotics and Automation Letters},
  year={2023},
  publisher={IEEE}
}

@inproceedings{neurad,
  title={Neurad: Neural rendering for autonomous driving},
  author={Tonderski, Adam and Lindstr{\"o}m, Carl and Hess, Georg and Ljungbergh, William and Svensson, Lennart and Petersson, Christoffer},
  booktitle={Proceedings of the IEEE/CVF Conference on Computer Vision and Pattern Recognition},
  pages={14895--14904},
  year={2024}
}

@inproceedings{unisim,
  title={Unisim: A neural closed-loop sensor simulator},
  author={Yang, Ze and Chen, Yun and Wang, Jingkang and Manivasagam, Sivabalan and Ma, Wei-Chiu and Yang, Anqi Joyce and Urtasun, Raquel},
  booktitle={Proceedings of the IEEE/CVF Conference on Computer Vision and Pattern Recognition},
  pages={1389--1399},
  year={2023}
}

@article{3dgs,
  title={3D Gaussian Splatting for Real-Time Radiance Field Rendering.},
  author={Kerbl, Bernhard and Kopanas, Georgios and Leimk{\"u}hler, Thomas and Drettakis, George},
  journal={ACM Trans. Graph.},
  volume={42},
  number={4},
  pages={139--1},
  year={2023}
}

@article{pvg,
  title={Periodic vibration gaussian: Dynamic urban scene reconstruction and real-time rendering},
  author={Chen, Yurui and Gu, Chun and Jiang, Junzhe and Zhu, Xiatian and Zhang, Li},
  journal={arXiv preprint arXiv:2311.18561},
  year={2023}
}

@article{streetgs,
  title={Street gaussians for modeling dynamic urban scenes},
  author={Yan, Yunzhi and Lin, Haotong and Zhou, Chenxu and Wang, Weijie and Sun, Haiyang and Zhan, Kun and Lang, Xianpeng and Zhou, Xiaowei and Peng, Sida},
  journal={arXiv preprint arXiv:2401.01339},
  year={2024}
}

@article{s3gs,
  title={S3 Gaussian: Self-Supervised Street Gaussians for Autonomous Driving},
  author={Huang, Nan and Wei, Xiaobao and Zheng, Wenzhao and An, Pengju and Lu, Ming and Zhan, Wei and Tomizuka, Masayoshi and Keutzer, Kurt and Zhang, Shanghang},
  journal={arXiv preprint arXiv:2405.20323},
  year={2024}
}

@inproceedings{drivinggs,
  title={Drivinggaussian: Composite gaussian splatting for surrounding dynamic autonomous driving scenes},
  author={Zhou, Xiaoyu and Lin, Zhiwei and Shan, Xiaojun and Wang, Yongtao and Sun, Deqing and Yang, Ming-Hsuan},
  booktitle={Proceedings of the IEEE/CVF Conference on Computer Vision and Pattern Recognition},
  pages={21634--21643},
  year={2024}
}

@article{tclcgs,
  title={Tclc-gs: Tightly coupled lidar-camera gaussian splatting for surrounding autonomous driving scenes},
  author={Zhao, Cheng and Sun, Su and Wang, Ruoyu and Guo, Yuliang and Wan, Jun-Jun and Huang, Zhou and Huang, Xinyu and Chen, Yingjie Victor and Ren, Liu},
  journal={arXiv preprint arXiv:2404.02410},
  year={2024}
}

@article{autosplat,
  title={Autosplat: Constrained gaussian splatting for autonomous driving scene reconstruction},
  author={Khan, Mustafa and Fazlali, Hamidreza and Sharma, Dhruv and Cao, Tongtong and Bai, Dongfeng and Ren, Yuan and Liu, Bingbing},
  journal={arXiv preprint arXiv:2407.02598},
  year={2024}
}

@article{krishnan2023lane,
  title={LANe: Lighting-aware neural fields for compositional scene synthesis},
  author={Krishnan, Akshay and Raj, Amit and Zhang, Xianling and Carlson, Alexandra and Tseng, Nathan and Sridhar, Sandhya and Jaipuria, Nikita and Hays, James},
  journal={arXiv preprint arXiv:2304.03280},
  year={2023}
}

@inproceedings{HUGS,
  title={Hugs: Holistic urban 3d scene understanding via gaussian splatting},
  author={Zhou, Hongyu and Shao, Jiahao and Xu, Lu and Bai, Dongfeng and Qiu, Weichao and Liu, Bingbing and Wang, Yue and Geiger, Andreas and Liao, Yiyi},
  booktitle={Proceedings of the IEEE/CVF Conference on Computer Vision and Pattern Recognition},
  pages={21336--21345},
  year={2024}
}

@article{liv-gaussmap,
  title={Liv-gaussmap: Lidar-inertial-visual fusion for real-time 3d radiance field map rendering},
  author={Hong, Sheng and He, Junjie and Zheng, Xinhu and Zheng, Chunran and Shen, Shaojie},
  journal={IEEE Robotics and Automation Letters},
  year={2024},
  publisher={IEEE}
}

@article{letsgo,
  title={LetsGo: Large-Scale Garage Modeling and Rendering via LiDAR-Assisted Gaussian Primitives},
  author={Cui, Jiadi and Cao, Junming and Zhong, Yuhui and Wang, Liao and Zhao, Fuqiang and Wang, Penghao and Chen, Yifan and He, Zhipeng and Xu, Lan and Shi, Yujiao and others},
  journal={arXiv preprint arXiv:2404.09748},
  year={2024}
}

@article{ligs,
  title={LI-GS: Gaussian Splatting with LiDAR Incorporated for Accurate Large-Scale Reconstruction},
  author={Jiang, Changjian and Gao, Ruilan and Shao, Kele and Wang, Yue and Xiong, Rong and Zhang, Yu},
  journal={arXiv preprint arXiv:2409.12899},
  year={2024}
}

@inproceedings{physgaussian,
  title={Physgaussian: Physics-integrated 3d gaussians for generative dynamics},
  author={Xie, Tianyi and Zong, Zeshun and Qiu, Yuxing and Li, Xuan and Feng, Yutao and Yang, Yin and Jiang, Chenfanfu},
  booktitle={Proceedings of the IEEE/CVF Conference on Computer Vision and Pattern Recognition},
  pages={4389--4398},
  year={2024}
}

@article{gsplat,
  title={gsplat: An open-source library for Gaussian splatting},
  author={Ye, Vickie and Li, Ruilong and Kerr, Justin and Turkulainen, Matias and Yi, Brent and Pan, Zhuoyang and Seiskari, Otto and Ye, Jianbo and Hu, Jeffrey and Tancik, Matthew and others},
  journal={arXiv preprint arXiv:2409.06765},
  year={2024}
}

@inproceedings{zhang2022simbar,
  title={Simbar: Single image-based scene relighting for effective data augmentation for automated driving vision tasks},
  author={Zhang, Xianling and Tseng, Nathan and Syed, Ameerah and Bhasin, Rohan and Jaipuria, Nikita},
  booktitle={Proceedings of the IEEE/CVF Conference on Computer Vision and Pattern Recognition},
  pages={3718--3728},
  year={2022}
}

@article{lidargs,
  title={LiDAR-GS: Real-time LiDAR Re-Simulation using Gaussian Splatting},
  author={Chen, Qifeng and Yang, Sheng and Du, Sicong and Tang, Tao and Chen, Peng and Huo, Yuchi},
  journal={arXiv preprint arXiv:2410.05111},
  year={2024}
}

@inproceedings{deformgs,
  title={Deformable 3d gaussians for high-fidelity monocular dynamic scene reconstruction},
  author={Yang, Ziyi and Gao, Xinyu and Zhou, Wen and Jiao, Shaohui and Zhang, Yuqing and Jin, Xiaogang},
  booktitle={Proceedings of the IEEE/CVF conference on computer vision and pattern recognition},
  pages={20331--20341},
  year={2024}
}

@inproceedings{pandaset,
  title={Pandaset: Advanced sensor suite dataset for autonomous driving},
  author={Xiao, Pengchuan and Shao, Zhenlei and Hao, Steven and Zhang, Zishuo and Chai, Xiaolin and Jiao, Judy and Li, Zesong and Wu, Jian and Sun, Kai and Jiang, Kun and others},
  booktitle={2021 IEEE international intelligent transportation systems conference (ITSC)},
  pages={3095--3101},
  year={2021},
  organization={IEEE}
}

@inproceedings{nuscenes,
  title={nuscenes: A multimodal dataset for autonomous driving},
  author={Caesar, Holger and Bankiti, Varun and Lang, Alex H and Vora, Sourabh and Liong, Venice Erin and Xu, Qiang and Krishnan, Anush and Pan, Yu and Baldan, Giancarlo and Beijbom, Oscar},
  booktitle={Proceedings of the IEEE/CVF conference on computer vision and pattern recognition},
  pages={11621--11631},
  year={2020}
}

@article{nerfgs_slam_survey,
  title={How nerfs and 3d gaussian splatting are reshaping slam: a survey},
  author={Tosi, Fabio and Zhang, Youmin and Gong, Ziren and Sandstr{\"o}m, Erik and Mattoccia, Stefano and Oswald, Martin R and Poggi, Matteo},
  journal={arXiv preprint arXiv:2402.13255},
  volume={4},
  pages={1},
  year={2024},
  publisher={Apr}
}

@article{stereo_gs_slam,
  title={Stereo 3D Gaussian Splatting SLAM for Outdoor Urban Scenes},
  author={Li, Xiaohan and Gong, Ziren and Tosi, Fabio and Poggi, Matteo and Mattoccia, Stefano and Liu, Dong and Wu, Jun},
  journal={arXiv preprint arXiv:2507.23677},
  year={2025}
}

\end{document}